\documentclass[final]{cvpr}

\usepackage{times}
\usepackage{epsfig}
\usepackage{graphicx}
\usepackage{amsmath}
\usepackage{amssymb}
\usepackage{multirow}
\usepackage{colortbl}
\usepackage{subcaption}
\usepackage{color}

\usepackage[pagebackref=true,breaklinks=true,colorlinks,bookmarks=false]{hyperref}

\def\cvprPaperID{1000} 
\def\confYear{CVPR 2021}
\begin{document}

\title{PGT: A Progressive Method for Training Models on Long Videos}

\author{Bo Pang\thanks{Equal contribution.} \quad\ Gao Peng\footnotemark[1] \quad\ Yizhuo Li \quad\ Cewu Lu\thanks{Cewu Lu is the corresponding author.}\\
	Shanghai Jiao Tong University\\
	\{pangbo, penggao, liyizhuo, lucewu\}@sjtu.edu.cn
}
\maketitle

\begin{abstract}
	\vspace{-0.13in}
   Convolutional video models have an order of magnitude larger computational complexity than their counterpart image-level models. Constrained by computational resources, there is no model or training method that can train long video sequences end-to-end. Currently, the main-stream method is to split a raw video into clips, leading to incomplete fragmentary temporal information flow. Inspired by natural language processing techniques dealing with long sentences, we propose to treat videos as serial fragments satisfying Markov property, and train it as a whole by progressively propagating information through the temporal dimension in multiple steps. This progressive training (PGT) method is able to train long videos end-to-end with limited resources and ensures the effective transmission of information. As a general and robust training method, we empirically demonstrate that it yields significant performance improvements on different models and datasets. As an illustrative example, the proposed method improves SlowOnly network by 3.7 mAP on Charades and 1.9 top-1 accuracy on Kinetics with negligible parameter and computation overhead. Code is available at: \href{https://github.com/BoPang1996/PGT}{https://github.com/BoPang1996/PGT}.

\end{abstract}

\vspace{-0.2in}
\section{Introduction}
\vspace{-0.07in}

  Semantic information often flows across a long time in videos. However, end-to-end modeling a long video as a whole is not feasible for current convolutional methods since their computational complexities linearly increase with the number of frames~\cite{tang2020asynchronous}. The main-stream solution is splitting a video into multiple short clips~\cite{tsn,nonlocal,slowfast,pang2020tubetk}, but in this way, video models can only access local fragmentary temporal information, thus, fail to model long semantics~\cite{lfb,tang2020asynchronous}. Is this trade-off between computational complexity and semantic integrity unavoidable, or might there be a specific training method tailored for video tasks that can model long semantics with acceptable complexity?
  
  \begin{figure}[t]
  	\begin{center}
  		\includegraphics[width=\linewidth]{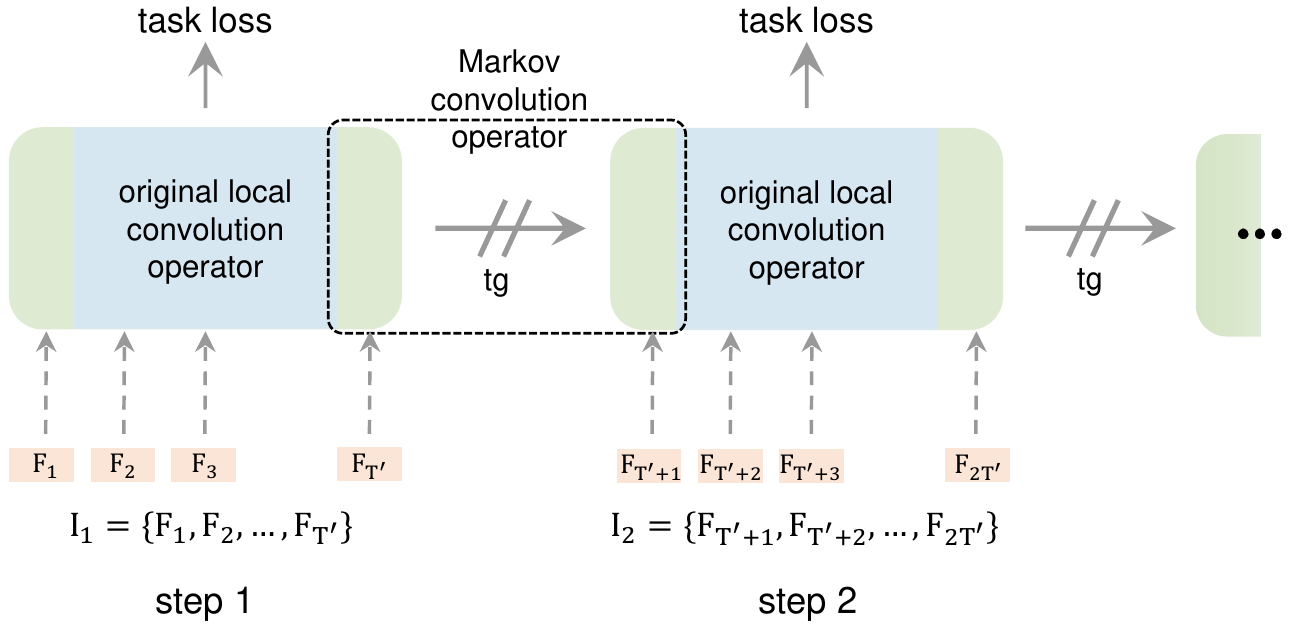}
  	\end{center}
  \vspace{-0.2in}
  	\caption{\textbf{Progressive training} (PGT) treats videos as serial fragments and optimizes a CNN model with multiple progressive steps on long videos. The Markov convolutional operator designed to transfer temporal features among steps is adopted on the first and last frames of each step, and the gradient is truncated between them. ``tg" denotes truncating gradients. $F_i$ is the $i$th video frame and $I_p$ is the input of the $p$th progressive step containing multiple frames.
  	}
  	\label{fig:cover}
  	\vspace{-0.2in}
  \end{figure}
  
  The main cause of this problem is that 3D convolutional models~\cite{c3d,i3d,r21d} treat a video signal $I(x, y, t)$ as an integrated information block and have to process it as a whole. With the video growing longer, the information block becomes larger and the processing complexity increases to an infeasible point. Since in the temporal dimension the development of video semantics has high-order Markov property, violently splitting long videos and processing short clips with convolutions to model local features will hurt semantic integrity. For example, the model will never know an action of ``pour milk" is making latte, unless it also sees the previous action of ``grinding coffee beans".
  
  To avoid the trade-off between computational complexity and semantic integrity --- \textit{i.e.}, to end-to-end train a model on long videos with much lower complexity, in this paper, we propose the progressive training (PGT) method (see Fig.~\ref{fig:cover}). 
  Inspired by Truncated Back-Propagation through Time (TBPTT)~\cite{tbptt} originally designed for recurrent neural networks to model long natural language sequences, the central idea of PGT is to 
  1) treat a video as serial fragments satisfying high-order Markov property instead of an integrated signal block, 
  2) disassemble the integrated forward and backward propagation into multiple serial portions like TBPTT (see \S\ref{sec:tbptt}), which doesn't break the Markov dependency of the calculation flow.
  Modeling a long video in multiple steps won't lead to high resource consumption and the Markov property ensures the integrity of temporal semantics after disassembling, akin to how TBPTT enables training RNN on long sequences.
  
\begin{figure*}[t]
	\begin{center}
		\includegraphics[width=\linewidth, height=2.0in]{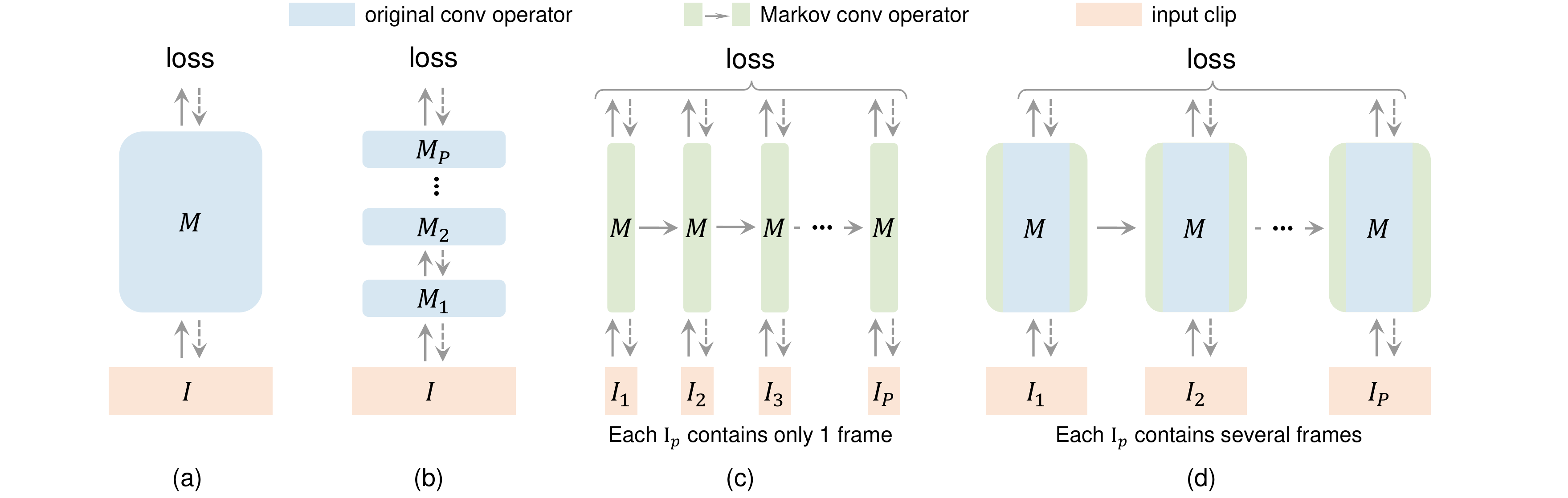}
	\end{center}
    \vspace{-0.2in}
	\caption{\textbf{Conceptual comparison of (a) baseline, (b) splitting model, and (c, d) splitting input progressive training}. The solid and dashed lines mean forward and backward paths. \textbf{(a)} The conventional training method trains the model $M$ with input $I$ in an integrated step. \textbf{(b)} In the splitting model setting, we split $M$ into several parts. Although these parts satisfy the one-way dependency and can be calculated progressively in forward-propagation, they break this one-way dependency in back-propagation, failing to achieve the serial progressive training. \textbf{(c)} The splitting input setting satisfies the one-way dependency constraint in both forward and backward propagation, thus, can be optimized step by step progressively. Here, only the Markov operator is adopted to transfer features among progressive steps. \textbf{(d)} This setting is similar to (c). Differences are that (d) has a larger progressive length for each step and within each step, besides Markov operators adopted at the edge of each step to transfer temporal information, original convolutional operators are adopted to capture internal temporal features.
	}
	\label{fig:schedule}
	\vspace{-0.2in}
\end{figure*}

  Because the common convolutional operator is a kind of local operator which does not satisfy the Markov property, we design several Markov convolutional operators with only a few modifications on the original convolutional operator so that they can easily replace the original one in modern video models when training. With these operators, the progressive training schedule mixing local and Markov features is proposed (see Fig.~\ref{fig:schedule} and Fig.~\ref{fig:markov_operator}): The temporal information is propagated progressively forward in multiple steps where within each progressive step, local operators capture current features together with those transferred from previous steps through the Markov operators.
  In this schedule, the Markov operators propagate temporal information among the progressive steps throughout the temporal dimension and the serial multi-step splitting reduces the computational resource requirements.
  
  The proposed PGT method is effective and pretty simple. It is easy to implement and typically requires small changes to a video model with negligible parameter or complexity overhead. Empirically, it works with default learning rate schedules and hyper-parameters already in use except for weight decay rates (longer inputs need stronger regularization). Extensive experiments show that the progressive training method works robustly out-of-the-box for different models (RegNet3D~\cite{regnet}, ResNet~\cite{resnet}, SlowFast~\cite{slowfast}), datasets (Kinetics-200~\cite{k200}, Kinetics-400~\cite{k400}, Charades~\cite{charades}, AVA~\cite{ava}), and training settings (e.g. from scratch or pre-trained). We observe consistent performance improvements without tuning. As an example, the progressive method improves SlowOnly network by 3.7 mAP on Charades and 1.9 top-1 accuracy on Kinetics. We hope this simple and effective method will provide the community with new insights into modeling long videos.

\vspace{-0.05in}
\section{Related Work}
\vspace{-0.1in}

\paragraph{Convolutional Video Networks}
Video convolutional backbones are developed from image-level backbone networks~\cite{resnet,alexnet,vgg,inception,tdaf}. Inchoate methods~\cite{twostream,feichtenhofer2016convolutional} directly apply image networks on optical-flow inputs~\cite{hall2019,piergiovanni2019representation} to model temporal information. Then researchers extend 2D convolutional operators to 3D ones by extending the temporal dimension~\cite{c3d,hara2018can} and 3D convolution based models~\cite{c3d,i3d,r21d,k200,karpathy2014large,gsta,assemblenet,vtn,p3d,sun2015human,lou2020human} (including ones adopting 2D spatial plus 1D temporal filters) become the mainstream method. Recently, non-local network~\cite{nonlocal} is designed to model global video features instead of local ones. A two-stream structure SlowFast~\cite{slowfast} is proposed to balance the spatial and temporal information. X3D~\cite{x3d} finds efficient video architectures by progressively extending each dimensions (e.g. spatial, temporal, channel dimensions). AssembleNet~\cite{assemblenet} proposes a method to automatically form connections among CNN blocks.

\vspace{-0.2in}
\paragraph{Methods to Handle Long Videos}
 How to use modern convolutional networks to handle long video tasks is less studied, due to limited computational resources and the behaviour of the convolutional operator. One simple method is to use large sampling strides to represent a long video with only a few frames~\cite{tsn,feichtenhofer2017spatiotemporal,tsm}, which causes serious information loss. A better strategy is to model long videos on the top of pre-computed deep features of short clips~\cite{li2017temporal,miech2017learnable,tang2018non,yue2015beyond,lfb,tang2020asynchronous} by pooling~\cite{tang2018non,miech2017learnable}, CNN~\cite{yue2015beyond,pic}, graph~\cite{videoG,poseflow}, memory blocks~\cite{tang2020asynchronous}, or attention methods~\cite{lfb}. However, because the features of short clips only contain local information and cannot be updated, these two-step methods without end-to-end training are likely suboptimal. Thus, we propose the first method to train deep CNN models end-to-end on long videos, requiring almost the same computational resources as short clips.

\vspace{-0.2in}
\paragraph{Sequential Methods for Video Tasks}
  Modeling videos as sequences is an alternative strategy for convolutional models. \cite{yue2015beyond,donahue2015long,li2018videolstm,sun2017lattice} adopt LSTM~\cite{lstm} layers (a kind of recurrent operator) to model video frame features generated by image-level CNN models. \cite{deeprnn,csc} tailor better recurrent layers that are easy to stack deep for higher-dimensional video information. These recurrent-based methods have advantages over convolutional ones for tasks sensitive to sequence order, such as video future prediction~\cite{oh2015action,villegas2017decomposing,xingjian2015convolutional}, trajectory prediction~\cite{sun2020recursive}, and video description~\cite{donahue2015long,venugopalan2014translating}. While for tasks that more focus on integrated features like action recognition~\cite{wu2015modeling,li2018videolstm,adha,i3d,slowfast,li2020pastanet,li2020detailed}, there is still a gap between the recurrent and convolutional models.
  
\vspace{-0.2in}
\paragraph{Truncated Back-Propagation through Time}\label{sec:tbptt}

  Back-Propagation Through Time, or BPTT~\cite{bptt}, is the training algorithm used to update weights in recurrent neural networks like LSTMs~\cite{lstm}. It unrolls input timesteps (frames for video), calculates and accumulates errors for each timestep. Then the network is rolled back up to update the weights. But, for a long sequence with length $n$, BPTT will consume lots of memory and the complexity will be extremely large. Truncated Back-Propagation through Time (TBPTT)~\cite{tbptt} solves this problem by periodically updating the network. It unrolls and forward propagates the input for $k_1$ steps, then backward propagates accumulated errors of the past $k_2$ unrolled steps to update the network ($k_1 \leq k_2 < n$). This process is repeated till unrolling all the inputs. Through splitting the integrated training process ($k_1=k_2 = n$) into several sub-processes, TBPTT reduces the resource requirement. More importantly, because of the Markov property of RNN, this splitting does not change the flowing path of temporal information.
  
  In this paper, inspired by TBPTT, we slightly modify the convolutional operator to satisfy the Markov property and design a progressive training method for long videos.

\vspace{-0.09in}
\section{Progressive Training for Video Models}
\vspace{-0.06in}
  Before introducing the proposed progressive training method, let's consider a reference convolutional video processing model (\textit{e.g.} C3D~\cite{c3d}, I3D~\cite{i3d}) that operates on videos of shape $T\times H \times W$ (number of frames $\times$ height $\times$ width). Due to the computational resource limitation, common methods are to split the raw video into short ones with length $T'<T$. This separation forces models to focus on short local temporal features, wasting the large receptive field of deep models (\textit{e.g.} SlowOnly with the receptive field of 39 only models 8 frames) and breaking the semantic integrity. Although there are methods for long videos \cite{miech2017learnable,tang2018non,yue2015beyond,lfb,tang2020asynchronous}, they design extra modules to model fixed features generated by backbone models which only take short clips as input, instead of end-to-end training the backbone and extra modules. Thus, the whole is likely suboptimal, since the backbone still models short local features.
  
  To end-to-end train long videos and ensure the complexity won't surpass the acceptable range, we propose the progressive training method. Inspired by Truncated Back-Propagation Through Time designed for recurrent methods, which reduces the complexity by separating each time stamp and truncating the computing graph of back-propagation, we realize that the end-to-end feature extraction and optimization process of a long video need not be finished in only one step. Instead, it can be a serial progressive process. We will show in experiments that this method can significantly enhance model's performance without introducing any parameter or complexity overhead. After analysis, we believe this improvement is mainly due to the increase of the temporal receptive field (see \S~\ref{sec:erf}).

\vspace{-0.07in}
\subsection{Progressive Method}
\vspace{-0.07in}
  The progressive training (PGT) method aims at disassembling an integrated computing process into several portions that can be calculated serially to reduce computing resources requirements. To make sure that the temporal semantics are not broken by the disassembling process, the equivalent disassembling is the best choice --- \textit{i.e.} the calculation flow is exactly the same before and after disassembling. To achieve this equivalency, the serial disassembling process needs to satisfy such a constraint: 
  
  \textbf{Constraint 1}: \textit{Among the split portions, there is only one-way dependency --- if the computing process of portion $A$ depends on results from portion $B$, then the computing of portion $B$ cannot depend on portion $A$.} 
  
  The disassemble progress satisfying this constraint can be formally expressed as:
  \vspace{-0.07in}
  \begin{equation}
  M(I) \Leftrightarrow M_P(I_P, M_{P-1}(I_{P-1}, M_{P-2}(I_{P-2}, ...))),
  \vspace{-0.07in}
  \end{equation}
  where $I = \{I_p|p\in\{1,2,...,P\}\}$ and $M = \{M_p|p \in \{1,...,P\}\}$ are whole input and model. $I_p$ and $M_p$ are small portions of them, $p$th in the progressive order. $P$ is the total number of split portions. This means that the disassembling can be achieved by splitting the input, splitting the model, or both of them. For example (see Fig.~\ref{fig:schedule}):
  
  \textbf{-} \textit{Splitting input}: we split an input video $I = \{F_1, F_2, ..., F_T\}$ with $T$ frames to several frame groups $I_p = \{F_{(p-1) \times T'+1},...,F_{p\times T'}\}$, where $T'$ is the length of each frame group. The computation of each group can only depend on the current and previous groups (here, model $M$ is not split, --- \textit{i.e.} $M = M_1 = M_2 = ... = M_P$);
  
  \textbf{-} \textit{Splitting model}: we split a deep model to several layer groups and compute group by group to the results (here, we do not split the input. Thus, $I_2 = I_3 = ... = I_P = {\rm None}$).
  
  Although with Constraint 1 we can split the forward propagation equivalently, progressively training a long video also needs to serially disassemble the back-propagation process by truncating the gradient among the portions to completely isolate them to reduce the training resource requirements. To make sure each portion can get gradients to update their parameters after truncating, the disassembling needs to satisfy another constraint:
   
  \textbf{Constraint 2}: \textit{The one-way dependency should be maintained in the back-propagation process ---
   Assume the computing of portion $A$ depends on portion $B$ and $B$ does not depend on $A$. $B$ should be able to update its parameters without gradients back-propagated from portion $A$.}
   
  Now let's go back to the ``\textit{Splitting model}" example, which doesn't satisfy this constraint (see Fig.~\ref{fig:schedule} (b)). Because fore layers can only get gradients from hind layers, if truncated, the fore layers will have no gradient to update their parameters. Thus, in the training phase, the split portions still need to be optimized together, failing to reduce the complexity of each portion by disassembling. Thus, in this paper, to satisfy both the two constraints, we only consider the ``splitting input" setting:
   \vspace{-0.07in}
  \begin{equation}~\label{eq: pgt}
    M(I) \Leftrightarrow M(I_P, M(I_{P-1}, M(I_{P-2}, ...))),
  \vspace{-0.07in}
  \end{equation}
  Eq.~\ref{eq: pgt} is similar to the calculation process of RNN, but our progressive method does not aim at building a RNN model, instead, providing a method to train any model (such as CNNs, transformers) satisfying the constraints in a progressive manner to reduce the training resource requirements.
  
  For an integrated input $I\in R^{T\times H \times W}$ with $T$ frames, we call the length $T'$ of $I_p\in R^{T' \times H \times W}$ as the ``progressive length" and $P$ as the ``number of progressive steps".

\vspace{-0.05in}
\subsection{Basic Progressive Training for Deep CNN}
\vspace{-0.07in}
  In this part, we will discuss how to apply the basic progressive training (PGT) method on convolutional networks.
  
  \textbf{Local operator} The conventional convolution is a local operator that models both the past and future information. Formally, one-dimensional temporal convolution operator can be expressed as:
  \vspace{-0.07in}
  \begin{equation}
  \rm Conv(f_{past}, f_{cur}, f_{future})
  \vspace{-0.07in}
  \end{equation} 
  where $\rm f_{past}, f_{cur}, f_{future}$ are features of past, current and future frames from the previous layer.
  In deep models, features of each frame relies on the intermediate results from far past and future frames (30 frames in I3D; 18 ones in SlowOnly) --- \textit{i.e.} the adjacent frames rely on each other. This violates the one-way dependency constraints, making it difficult to adopt PGT to serialize the computation.
  
  \textbf{Basic Markov operator}~To solve the problem, we slightly modify the original local temporal convolutional operator to Markov one --- \textit{i.e.} hind frames depend on fore frames, while fore ones do not depend on hind ones in both forward and backward processes (Constraint 1 \& 2). The basic Markov convolutional operator (MCO) is to replace $\rm f_{future}$ with zero-padding in forward-propagation and truncate gradients to prevent it back-propagating through time:
  \vspace{-0.07in}
  \begin{equation}
  \begin{aligned}
   & \rm f_{res} = Conv(f_{past}, f_{cur}, 0)\\
   & \rm \frac{\partial{f_{res}}}{\partial{f_{past}}} := 0
  \end{aligned}
  \vspace{-0.07in}
  \end{equation}
  
  This simple modification does not aim at improving the original model. It is a compromise to make the deep CNN model satisfy the one-way dependency in Constraint 1 \& 2. We can apply the most fine-grained progressive training method (see Fig.~\ref{fig:schedule} (c)) with it: every progressive step only propagates information forward and backward one frame --- \textit{i.e.} the progressive length $T'= 1$ and the number of progressive steps $P=T$.
  
\begin{figure}[t]
 	\begin{center}
  		\includegraphics[width=\linewidth]{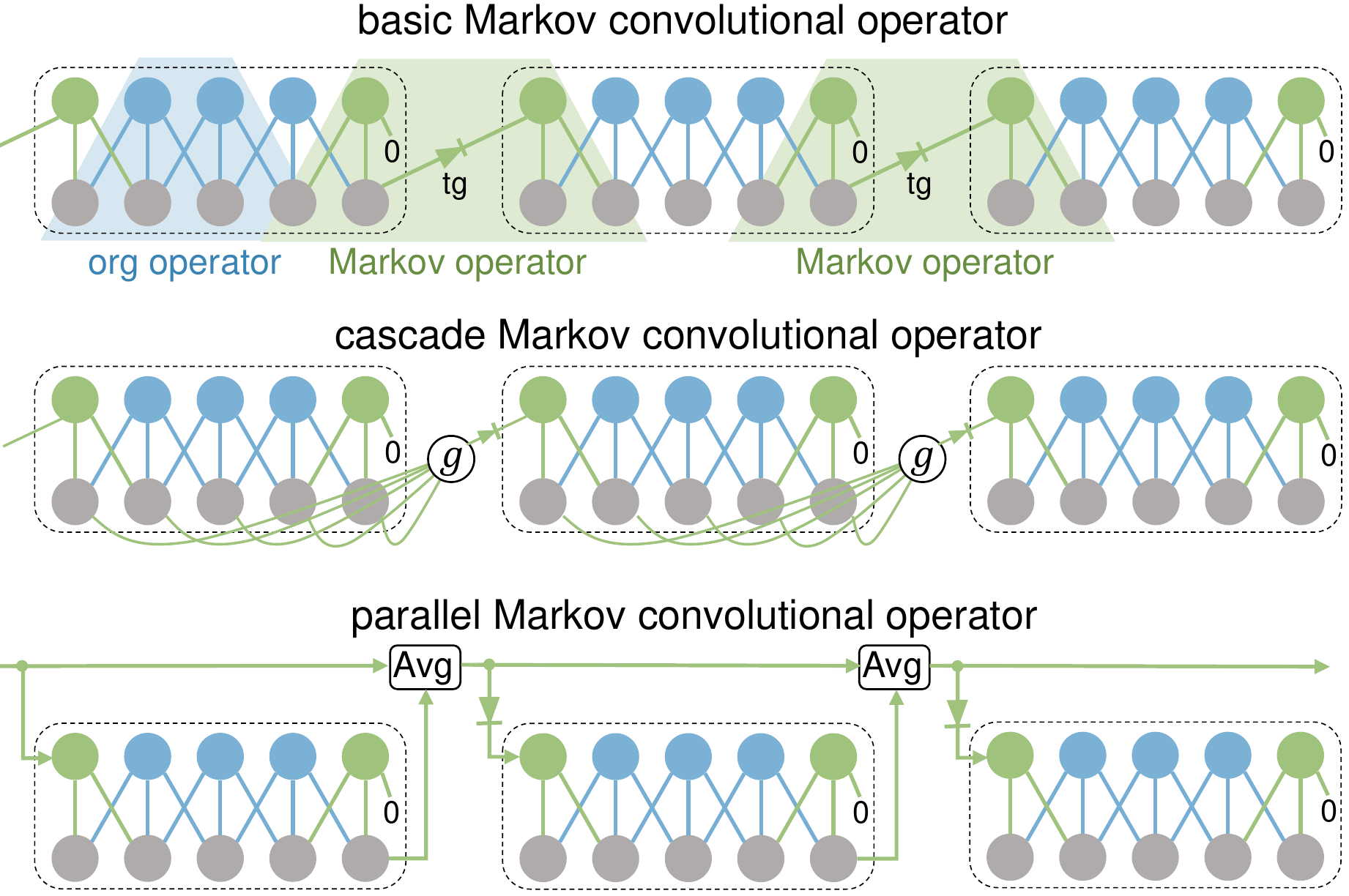}
 	\end{center}
 \vspace{-0.2in}
 	\caption{\textbf{Illustration of the proposed Markov convolutional operators}. Compared with the original convolutional operator, the Markov convolutional operator (MCO) satisfies the one-way dependency constraint. Cascade MCO enhances the temporal information flow between two progressive steps. Parallel MCO intensifies long-term features transfer. ``tg" denotes ``truncate gradient". 	}
 	\label{fig:markov_operator}
 	 \vspace{-0.2in}
\end{figure}

  \textbf{Basic progressive schedule} The ``Fine-grained" progressive method above is not necessary, because it makes the computing complexity of each step too low to take advantage of all the hardware's parallel computing power. More importantly, compared with the original local convolutional operator, the basic Markov one has a worse ability to extract temporal features since it weakens the temporal information flows. To this end, we set the progress length $T'$ to the same as the ordinary length of short clips generated by conventional preprocessing (\textit{e.g.} 8 or 32), and meanwhile use the original and Markov convolutional operators in combination. Specifically, we adopt modified operators on the first and last frames of each progressive step, where the beginning one truncates the gradients and the ending one zero-pads $\rm f_{future}$, to satisfy the one-way dependency and make Markov property exist between them. And for the frames in the middle, the original operator is utilized (see Fig.~\ref{fig:schedule}d and Fig.~\ref{fig:markov_operator}). Note that the original and the modified Markov operators share the same parameters.
  
  With these basic Markov operator and progressive schedule, we apply the progressive training method on CNN models and we call it our \textit{basic progressive training method} (Fig.~\ref{fig:markov_operator}). It's worth noting that in this basic method, the modification is tiny and can be applied on any convolutional network. In the following parts, we develop more delicate Markov operators (\S\ref{sec: markov operator}) and progressive schedules (\S\ref{sec: progressive schedule}) to get our full progressive training method.
  
\vspace{-0.03in}
\subsection{Further Designs on Markov operators}
\vspace{-0.06in}
\label{sec: markov operator}
  When designing Markov convolutional operators, we make sure that they won't introduce more parameters or heavy computation overhead. A further criterion is that operators should strengthen the efficiency of temporal information propagation, unlike the basic Markov convolutional operator which only restricts the information transfer to satisfy the two constraints. We experiment with the following two advanced operators (see Fig.~\ref{fig:markov_operator}):
  
  \noindent
  \textbf{Cascade Markov convolutional operator (CMCO)} In basic Markov operator, $\rm f_{past}$ is just features of the adjacent previous frame. To better propagate the integrated feature of the last progressive step, CMCO sets $\rm f_{past}$ as the aggregation of features from all the frames in the previous progressive step. The aggregation function $g(\cdot)$ can be average/max pooling or other reasonable choices. CMCO builds denser information paths to enhance feature propagation.

\noindent
  \textbf{Parallel Markov convolutional operator (PMCO)} In CMCO, we enhance the connection between the current and adjacent previous progress steps. In PMCO, we intensify the information flows  from all the previous steps. Specifically, we set $\rm f_{past}$ as the momentum average of the features from all the previous progressive steps. PMCO allows features to transfer better over a long time horizon.

  These operators do not change the core computation of the basic Markov convolutional operator, just modify the input features by adding some preprocessing. Note that the preprocessing only contains a few addition operations and the overhead can be ignored. 

\vspace{-0.1in}
\subsection{Further Designs on Progressive Schedule}
\vspace{-0.07in}
\label{sec: progressive schedule}
 Here, we further refine the progressive schedule. In the basic version, the progressive length $T'$ and number of progressive step $P$ are fixed values. Thus, the positions of the local and Markov operators in a long sequence are fixed too. Because the behaviour of the local and Markov operators is not the same, as the Markov one has a relatively weaker feature extraction ability than the local operator, the temporal feature capture will be uneven, leading to information propagation bottlenecks.
 
 To avoid above problems caused by such unevenness, we propose the dynamic progressive regularization (DPR). Specifically, we randomly jitter the value of $T'$ and $P$ around the values of basic schedule and keep $T'\times P$ floating around a constant value. This method adjusts the ratio of local operators to Markov operators and meanwhile makes the position of Markov operators more evenly distributed in the long video to reduce the effect caused by information propagation bottlenecks. From another point of view, similar to Dropout~\cite{dropout}, DPR which randomly adjust the paths of forward and backward propagation is kind of a model regularization. Thus, we call it dynamic progressive regularization. 
 In experiments, we adopt two DPR settings:
 
 \textbf{-} \textit{DPR-A}: We randomly choose $T'$ from the set $\{0.75T'_b, T'_b, 1.25T'_b\}$ where $T'_b$ is the progressive length of the basic schedule. Note that when the progressive length is $1.25T'_b$, the training complexity of each step will be higher.
 
 \textbf{-} \textit{DPR-B}: We randomly choose $T'$ from the set $\{0.5T'_b, 0.75T'_b, T'_b\}$. This setting adopts more Markov operators and won't cause complexity increase.

\vspace{-0.1in}
\subsection{Implementation Details}
\vspace{-0.07in}
  \noindent\textbf{Optimizer}~
  We choose SGD as our optimizer and its specific settings like momentum and learning rate schedule are kept the same with corresponding baselines. For different tasks and datasets, optimization details are given in their experiment sections. Different from the conventional training method, when training with PGT, we accumulate the gradients of each progressive step and update parameters after the backward propagations of all the progressive steps.

  \noindent\textbf{Progressive implementation}~
  We keep the same frame sampling stride as baselines. For example, a common sampling method is $T\times \tau = 8\times8$, where $T$ and $\tau$ are the number of sampling frames and the sampling stride. We keep the stride $\tau$ and only enlarge $T$ to train a long video instead of short clips. Considering the video length in commonly used datasets, we set the number of progressive step $P=5$ for the basic schedule, unless otherwise specified. We let the two adjacent progressive steps overlap by one frame for better performance. Thus, the total frame in progressive input is $T = (T' - 1) * P + 1$. For DPR, given the total number of input frames $T_b$ of basic schedule, the number of progressive step $P$ is calculated as $P = {\rm round}[(T_b-1)/(T'-1)]$ to keep the total length basically the same.

  \noindent\textbf{Inference method}~\label{sec:inference method}~
  Like training, the mainstream inference method also splits long videos into short clips, tests them separately, and averages the outputs to get final results. This method needs to test multiple views and usually, there are overlaps among the views. The proposed progressive training method trains a long video end-to-end, thus, when inference, we still test a long video directly. Although its inference complexity of one view is higher, we can adopt much fewer views to cover the whole video. We experiment with two inference methods:
  
  	\textbf{-} \textit{Original long view} (orig long): In this mode, we get rid of the Markov operators and only utilize the original convolution for all the frames in the long video. It is the same as testing a long video in only one progressive step.
  	
  	\textbf{-} \textit{Progressive long view} (PG long): Just like the training phase, a long video is tested in multiple progressive steps.

\vspace{-0.1in}
\section{Experiments on Kinetics}
\vspace{-0.05in}
  We first evaluate our progressive training method on the large scale action recognition benchmark Kinetics-400~\cite{k400} and provide ablations on its subset Mini-Kinetics-200~\cite{k200}.
  
  \textbf{Dataset}
  Kinetics-400 contains 240k training and 20k validation videos covering 400 action categories. Its subset Mini-Kinetics-200 includes 200 categories and each one contains 400 training samples and 25 validation samples, resulting in 80k training and 5k validation samples in total.

  \textbf{Training}
  We adopt ResNet-3D (SlowOnly)~\cite{slowfast}, RegNet-3D~\cite{regnet}, and SlowFast~\cite{slowfast} as our baselines. The baseline training recipe follows \cite{slowfast}. We run SGD for 196 epochs on 16 GPUs with a mini-batch of 8 clips per GPU with initial learning rate of 0.2. The half-period cosine learning rate schedule~\cite{loshchilov2016sgdr} is adopted. We use a weight decay of $10^{-4}$, momentum of 0.9, and a linear learning rate warm-up~\cite{goyal2017accurate} from 0.02 over 34 epochs. Video clips are resized with shorter side $\in[256, 320]$ and inputs are randomly cropped patches with size of $224\times224$. Temporal sampling method is $T\times \tau = 8\times 8$. For progressive training, we enlarge the weight decay to $2\times 10^{-4}$, set $T'=8$ and $P = 5$. The total training epoch reduces to 100 and other recipes keep the same with baselines.
 
  \textbf{Inference} 
  Following~\cite{slowfast}, 10 clips with size $T\times \tau=8\times8$ are uniformly sampled from a video along temporal dimension. Each frame is resized with shorter side of 256 pixels and three patches of size $256\times256$ are taken to cover the spatial dimensions. Thus, there are $10\times 3$ views in total for baselines. For PGT, each clip has a size of $T'\times P \times\tau = 8\times5\times8$, covering a much longer range. Thus, we only take 2 temporal views ($2\times3$ views in total).

\vspace{-0.07in}
\subsection{Ablation Study}
\vspace{-0.07in}
  This section provides ablation studies on Mini-Kinetics-200 comparing accuracy and computational complexity. For convenient comparison, we express complexity as one-view Flops $\times$ progressive steps $P \times$ number of views $v$.  

\begin{table}[t]
	
	\caption{\textbf{PGT's performance on Mini-Kinetics-200}. We express the inference complexity as single-view GFLOPs $\times$ number of progressive steps $P \times$ number of views $v$. ``full PGT" adopts PMCO and DPR-B.}
	\vspace{-0.1in}
	\renewcommand{\arraystretch}{1.0}
	\centering
	\footnotesize
	\setlength\arrayrulewidth{0.7pt}
	\resizebox{\columnwidth}{!}{
	\begin{tabular}{ l|c  c | c}
		model & top-1 & top-5 & GFLOPs $\times P \times$ v\\
		\hline
		\rowcolor[gray]{0.95} RegNet0.4G-3D & 74.5 & 92.3 & 6.59 $\times$ 1 $\times$ 30\\
		\rowcolor[gray]{0.95} \textbf{RegNet0.4G-3D, + basic PGT} & 76.6 & 93.0 & 6.59 $\times$ 5 $\times$ 6\\
		\rowcolor[gray]{0.95} \textbf{RegNet0.4G-3D, + full PGT} & 77.5 & 93.6 & 6.59 $\times$ 5 $\times$ 6\\
		
		SlowOnly, R50 & 77.1 & 93.4 & 54.5 $\times$ 1 $\times$ 30\\
		\textbf{SlowOnly, R50, + basic PGT} & 78.9 & 94.0 & 54.5 $\times$ 5 $\times$ 6\\
		\textbf{SlowOnly, R50, + full PGT} & 79.6 & 94.2 & 54.5 $\times$ 5 $\times$ 6\\
	\end{tabular}}
	\label{tab:k200_res}  
	\vspace{-0.2in}
\end{table}

  \textbf{Basic and full progressive training}
  In Tab.~\ref{tab:k200_res}, we report the performance comparison between the progressive training (PGT) method and baselines to preliminarily reveal PGT's effectiveness. Since PGT does not modify models' overall structure, the one-view complexities of PGT are the same as the baselines. PGT processes 5 times longer videos than the baseline. Thus, it needs fewer temporal views to cover the whole video. Their total complexities are (almost) the same --- \textit{i.e.} PGT does not involve any overhead. From Tab.~\ref{tab:k200_res}, it is seen that PGT improves the top-1 accuracy of RegNet0.4G~\cite{regnet} by 3.0\% and SlowOnly-50~\cite{slowfast} by 2.5\%, revealing the importance of intact global temporal features that PGT focuses on.
  
  Next, Tab.~\ref{tab:abl} shows a series of ablations on the progressive training designs and inference methods, mainly using the RegNet0.4G-3D network, analyzed in turn.
  
\begin{table*}
	\caption{\textbf{Ablations on Mini-Kinetics-200}. Experiments are mainly based on RegNet0.4G-3D. For baseline, $T\times \tau=8\times8$. $P=5$ when adopting PGT. } 
	\vspace{-0.1in}
	\renewcommand{\arraystretch}{1.0}
	\centering
	\footnotesize
	\setlength\arrayrulewidth{0.7pt}
	\begin{subtable}[t]{0.33\linewidth}
		\centering
		\scriptsize
		\caption{\textbf{Markov Operators} Operators enhanced temporal information flow achieve better performances than the basic one.}
		\vspace{-0.05in}
		\begin{tabular}{ l| l |c  c }~\label{tab:abl-markov}
			model & operator & top-1 & top-5 \\
			\hline
			\rowcolor[gray]{0.95} RegNet0.4G-3D & baseline & 74.5 & 92.3\\
			\rowcolor[gray]{0.95} RegNet0.4G-3D & basic MCO & 76.6 & 93.0\\
			RegNet0.4G-3D & CMCO-avg & 77.1 & 93.3\\
			RegNet0.4G-3D & CMCO-max & 76.9 & 93.2\\
			RegNet0.4G-3D & PMCO & 77.3	& 93.3 \\
			\hline
			\rowcolor[gray]{0.95} SlowOnly, R50 & baseline & 77.1 & 93.4 \\
			\rowcolor[gray]{0.95} SlowOnly, R50 & basic MCO & 78.9 & 94.0 \\
			SlowOnly, R50 & PMCO & 79.3 & 94.1\\
		\end{tabular}
	\end{subtable}
	~~~~
	\begin{subtable}[t]{0.25\linewidth}
		\centering
		\scriptsize
		\caption{\textbf{PGT Schedules} More progressive steps perform better and DPR is more effective for more steps.}
		\vspace{-0.05in}
		\label{tab:abl-schedule}
		\begin{tabular}{ l| l |c  c }
			$P$ & schedule & top-1 & top-5 \\
			\hline
			1 step & baseline & 74.5 & 92.3\\
			\hline
			\rowcolor[gray]{0.95} 5 step & basic & 76.6 & 93.0 \\
			\rowcolor[gray]{0.95} 5 step & DPR-A & 77.3 & 93.4\\
			\rowcolor[gray]{0.95} 5 step & DPR-B & 77.2 & 93.4\\
			4 step & basic & 76.3 & 92.8 \\
			4 step & DPR-B & 76.7 & 93.0\\
			\rowcolor[gray]{0.95} 3 step & basic & 75.8 & 92.7\\
			2 step & basic & 75.1 & 92.6\\
		\end{tabular}
	\end{subtable}
	~~~~
	\begin{subtable}[t]{0.32\linewidth}
		\centering
		\scriptsize
		\caption{\textbf{Inference Methods} Performances of the baselines and progressive training methods with different inference schemes.}
		\vspace{-0.05in}
		\label{tab:abl-test}
		\begin{tabular}{ l| l |c  c }
			train method & test method & top-1 & top-5 \\
			\hline
			\rowcolor[gray]{0.95} baseline & 10$\times$3 short clip & 74.5 & 92.3\\
			\rowcolor[gray]{0.95} baseline & 1$\times$3 orig long & 74.7 & 92.4\\
			basic PGT & 10$\times$3 short clip & 75.8 & 92.7\\
			basic PGT & 1$\times$3 orig long & 76.1 & 92.7 \\
			basic PGT & 1$\times$3 PG long & 75.6 & 92.7\\
			basic PGT & 2$\times$3 orig long &76.6 & 93.0 \\
		    \rowcolor[gray]{0.95} full PGT & 2 $\times$ 3 orig long & 77.5 & 93.6\\
			\rowcolor[gray]{0.95} full PGT & 2 $\times$ 3 PG long & 76.7 & 93.1\\
		\end{tabular}
	\end{subtable}
	\label{tab:abl}
	\vspace{-0.15in}
\end{table*}

  \textbf{Markov operators}
  Tab.~\ref{tab:abl-markov} shows the performance of various Markov convolutional operators. As a naive Markov operator, the basic MCO improves the performance significantly as the previous paragraph states, although it sacrifices some feature extraction capabilities to transfer semantics among progressive steps. CMCO and PMCO alleviate this weakness to some extent and further improve the performance by 0.5\% and 0.7\%. For the following experiments (except ablations), we employ PMCO as our default.

  \textbf{PGT schedules}~
  Different feature extraction capacities of the Markov and local convolutional operators make temporal features uneven. We solve this problem by introducing the dynamic progressive regularization (DPR). Tab.~\ref{tab:abl-schedule} shows the effectiveness of it. It is seen that more progressive steps lead to better performance. DPR-A and DPR-B have almost the same improvements, where DPR-A is only a little bit ahead. Since DPR-B has lower complexity, we employ it as our default for following experiments.
  
  \textbf{Inference methods}~
  For PGT, we test the performance with longer clips and fewer views. In Tab.~\ref{tab:abl-test}, we also test the baseline model with this setting (first two lines) and we can see that it achieves similar performance to the multi-view short-clip setting, without substantial improvements. This reveals that it is the progressive training process that makes the model extract better temporal features instead of the longer test setting. The same conclusion can be drawn by comparing the 1st \& 3rd or 2nd \& 4th rows.
  
  Then we compare the two inference methods designed for PGT mentioned in \S\ref{sec:inference method}. It is seen that the \textit{original long view} inference method performs better on Kinetics. In the next section, we will show that the \textit{progressive long view} method is more suitable for Charades which contains longer activities consisting of several sub-actions.

\vspace{-0.07in}
\subsection{Main Results}
\vspace{-0.07in}
Tab.~\ref{tab:k400_res} shows the comparison with the SOTA results on Kinetics-400 for PGT method with different backbones: RegNet~\cite{regnet}, ResNet~\cite{resnet}, SlowFast~\cite{slowfast}, and Nonlocal~\cite{nonlocal}. 

In comparison to the advanced video baselines, our PGT method consistently provides a performance boost with negligible complexity overhead. For single-stream models such as RegNet and ResNet, PGT improves the performance by $\sim$1.8 accuracy. As for SlowFast with two streams, PGT provides 1.0\% improvements, which is higher than adopting Nonlocal with 10\%$\sim$20\% complexity overheads.

\begin{figure}[t]
	\begin{center}
		\includegraphics[width=\linewidth]{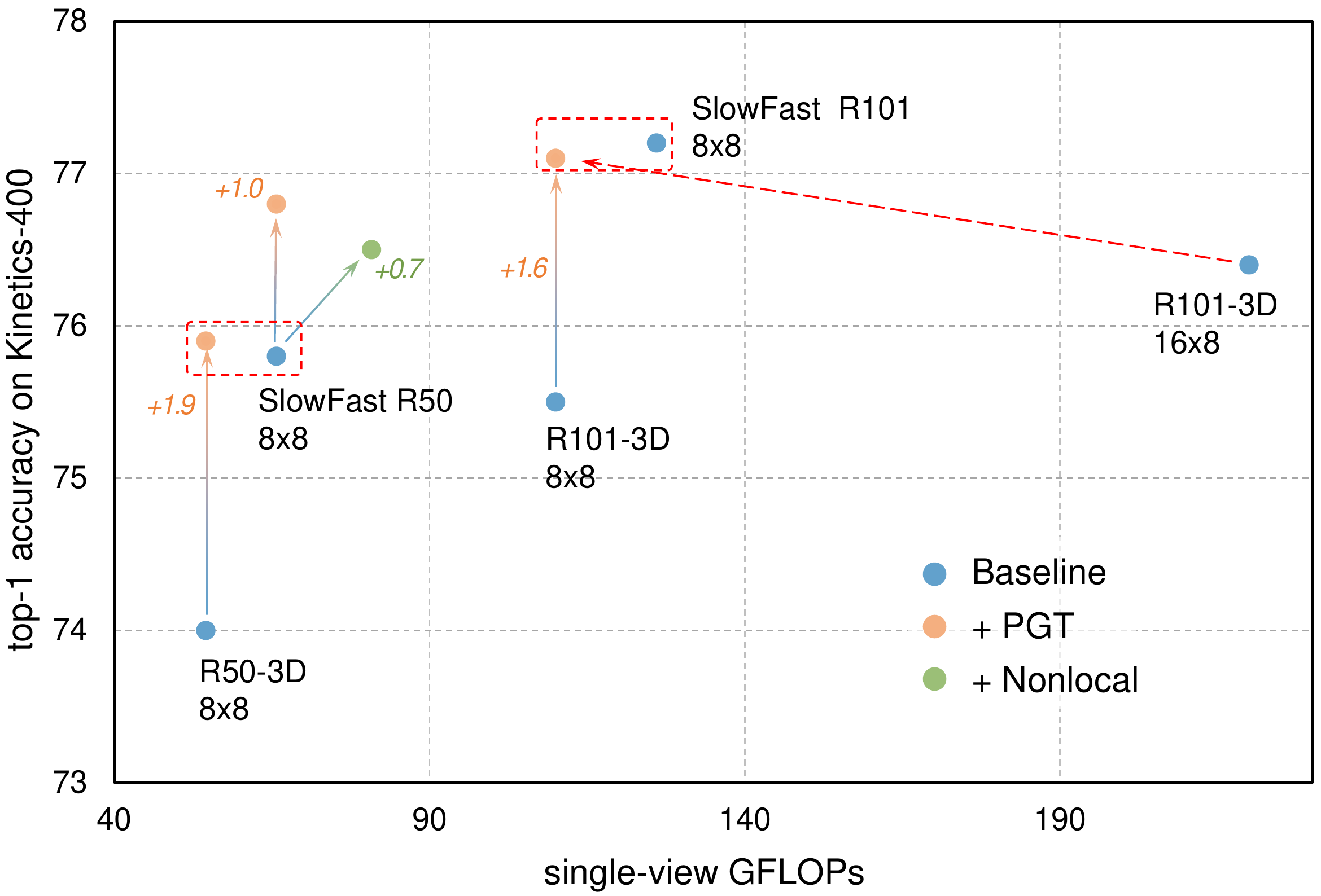}
	\end{center}
    \vspace{-0.2in}
	\caption{\textbf{Accuracy \& complexity comparison} on Kinetics-400. PGT method achieves consistent improvements on all the baselines with negligible overhead. As the red dashed line and red boxes shows, the PGT method achieves a better performance/complexity trade-off.
	}
	\label{fig:k400-compare}
	\vspace{-0.25in}
\end{figure}

Comparisons of the performance and complexity trade-off are shown in Fig.~\ref{fig:k400-compare}. The horizontal axis measures the single-step single-view GFLOPs with 256$\times$256 pixels input. We can see that PGT achieves better performance with lower complexity as the red arrow shows. The red boxes show that PGT leads to almost the same improvements as SlowFast does with lower complexities.

\begin{table}[t]
	\caption{\textbf{Comparison with the SOTAs on Kinetics-400.} ``flow" column indicates whether to adopt the optical flow and ``preT" denotes pre-trained on ImageNet.}
	\vspace{-0.1in}
	\renewcommand{\arraystretch}{1.0}
	\centering
	\footnotesize
	\setlength\arrayrulewidth{0.7pt}
	\resizebox{\columnwidth}{!}{
	\begin{tabular}{ l | c  c | c c | c }
		& & & & & inference \\
		model & flow & preT & top-1 & top-5 & {\scriptsize GFLOPs$\times P \times v$}\\
		
		\hline
		I3D~\cite{i3d} & & \checkmark & 72.1 & 90.3 & 108$\times$1$\times$N/A\\
		Two-Stream I3D & \checkmark & \checkmark & 75.7 & 92.0 & 216$\times$1$\times$N/A\\
		S3D-G~\cite{k200} & \checkmark & \checkmark & 77.2 & 93.0 & 143$\times$1$\times$N/A \\
		Nonlocal R50~\cite{nonlocal} & & \checkmark & 76.5 & 92.6 & 282$\times$1$\times$30 \\
		Nonlocal R101 & & \checkmark & 77.7 & 93.3 & 359$\times$1$\times$30 \\	
		\hline
		STC~\cite{stc} &  &  & 68.7 & 88.5 & N/A$\times$1$\times$N/A \\
		ARTNet~\cite{artnet} & &  & 69.2 & 88.3 & 23.5$\times$1$\times$250\\
		S3D~\cite{k200} & &  & 69.4 & 89.1 & 66.4$\times$1$\times$N/A \\
		ECO~\cite{eco} & &  & 70.0 & 89.4 & N/A$\times$1$\times$N/A \\
		I3D~\cite{i3d} & \checkmark &  & 71.6 & 90.0 & 216$\times$1$\times$N/A \\
		R(2+1)D~\cite{r21d} & \checkmark & & 73.9 & 90.9 & 304$\times$1$\times$115\\
		X3D-M~\cite{x3d} & &  & 76.0 & 92.3 & 6.2$\times$1$\times$30 \\
		X3D-XL~\cite{x3d} & &  & 79.1 & 93.9  & 48.4$\times$1$\times$30\\
		\hline
		\rowcolor[gray]{0.95} RegNet0.4G-3D~\cite{regnet} & &  & 70.3 & 89.3 & 6.59$\times$1$\times$30\\
		\rowcolor[gray]{0.95} \textbf{RegNet0.4G-3D + PGT} &  &  & 72.1 & 90.7 & 6.59$\times$5$\times$6\\
		R50-3D~\cite{slowfast} & & & 74.0 & 91.3 & 54.5$\times$1$\times$30\\
		\textbf{R50-3D ~~~~~~~~~~~~~~+ PGT} & &  & 75.9 & 92.4& 54.5$\times$5$\times$6\\
		\rowcolor[gray]{0.95} R101-3D~\cite{slowfast} & &  & 75.5 & 91.9 & 110$\times$1$\times$30\\
		\rowcolor[gray]{0.95} \textbf{R101-3D ~~~~~~~~~~~~+ PGT} & & & 77.1 & 92.9 & 110$\times$5$\times$6\\
		SlowFast R50~\cite{slowfast} & &  & 75.8 & 92.0 & 65.7$\times$1$\times$30 \\
		SlowFast R50 ~~~~~+ NL & &  & 76.5 & 92.4  & 80.8$\times$1$\times$30\\
		\textbf{SlowFast R50 ~~~~+ PGT} & &  & 76.8 & 92.6 & 65.7$\times$ 5$\times$6 \\
		\rowcolor[gray]{0.95} SlowFast R101~\cite{slowfast} & &  & 77.2 & 92.8 & 126$\times$1$\times$30 \\
	\end{tabular}}
	\label{tab:k400_res}  
	\vspace{-0.2in}
\end{table}  

\vspace{-0.07in}
\subsection{Analysis on Receptive Field}
\vspace{-0.07in}
\label{sec:erf}
  Modern deep convolutional video models often have large theoretical temporal receptive fields, such as 39 for SlowOnly network. Theoretical receptive field (TRF) is the upper bound of effective receptive field (ERF)~\cite{erf} and from \cite{erf}, we know the ERF will gradually increase during the training process, however, the input short clip generated by the mainstream temporal cropping training method is much shorter than the TRF, making model's ERF difficult to achieve a relative larger value after training.

\begin{figure}[t]
 	\begin{center}
 		\includegraphics[width=\linewidth, height=2.5in]{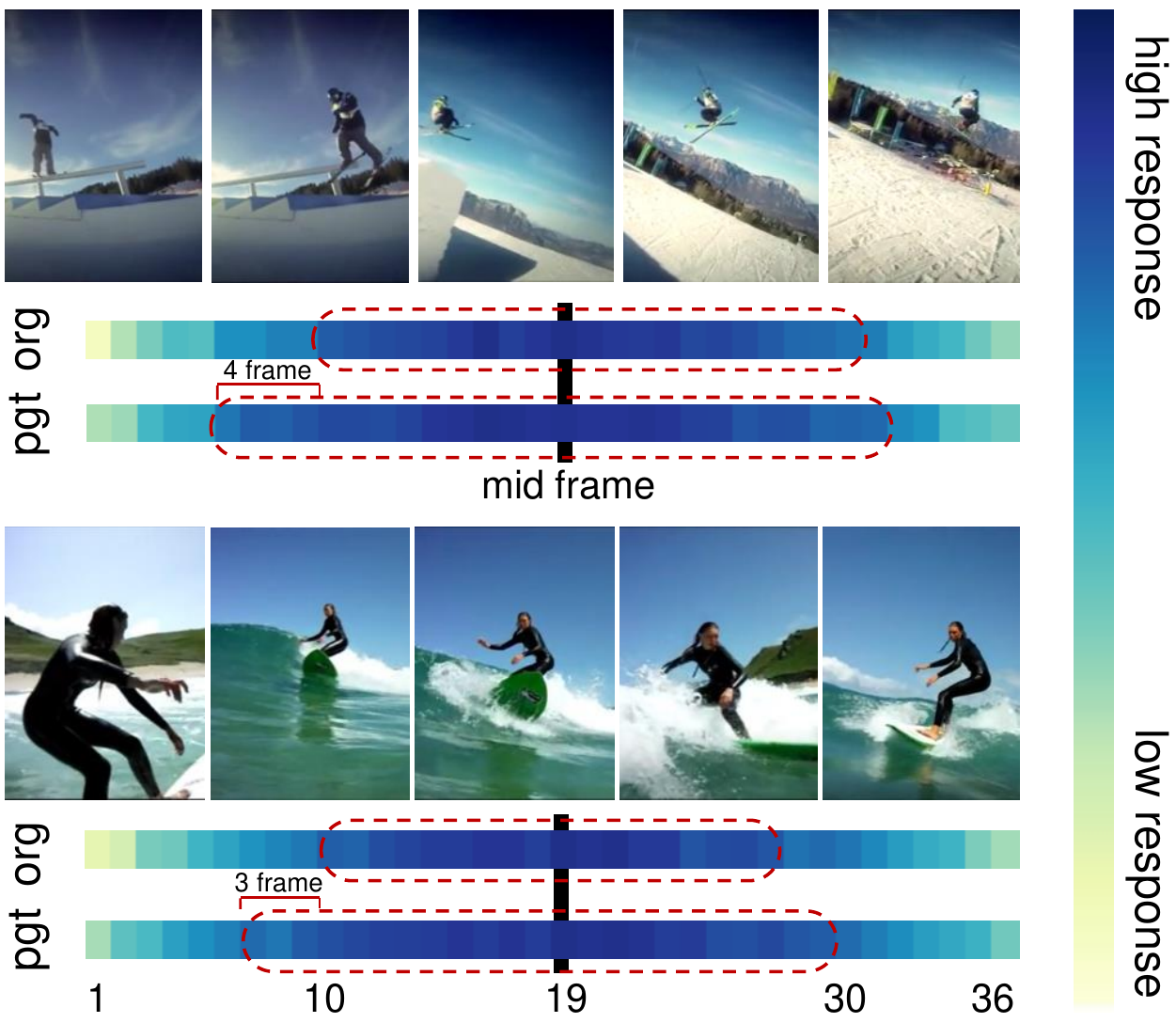}
 	\end{center}
    \vspace{-0.25in}
 	\caption{\textbf{Effective receptive field} (ERF) of original (org) and progressive (pgt) training method. The color bars show the ERF of the 19th frame with 36 input frames in total. From the red dashed boxes we can see that the progressive training method has a 40\% larger ERF on previous frames.
 	}
 	\label{fig:erf}
 	\vspace{-0.25in}
\end{figure}

  Here we conduct experiments to reveal that the proposed progressive training method can alleviate this problem. We adopt the SlowOnly network and train it on the Kinetics-200 dataset~\cite{k200}. Qualitative results are shown in Fig.~\ref{fig:erf}. We can see that compared with the original method trained on 8 frames, our progressive training method with $T'=8$ and $P=5$ has a 40\% larger ERF.
  
\vspace{-0.1in}
\section{Experiments on Charades}
\vspace{-0.07in}
  We then evaluate the proposed method on Charades dataset~\cite{charades} containing longer range activities spanning 30s on average. It contains about 9.8k training and 1.8k validation videos in 157 action categories (multi-labelled).
  
\textbf{Implementation details}~
We adopt SlowFast~\cite{slowfast} and ResNet-3D (SlowOnly)~\cite{slowfast} as our baselines. The models are pre-trained on Kinetics-400 or Kinetics-600~\cite{k600} following prior works~\cite{slowfast,x3d}.  
For PGT, we set $T'= 16$ and $P = 5$ for both training and inference. PMCO and CMCO-max are adopted together here.
We adopt ``pg long" inference method and get the final results with max-pooling over frames, instead of average-pooling since Charades is a multi-classification dataset.

\textbf{Results}~
Charades contains longer range activities. Thus, it can better reflect the advantages of PGT designed for long videos. Tab.~\ref{tab:charades_res} shows the results with ResNet~\cite{resnet}, SlowFast~\cite{slowfast}, and Nonlocal~\cite{nonlocal} as backbones. It is seen that our PGT consistently provides a performance boost: \textbf{3.1} mAP improvements on average for SlowOnly and SlowFast. It is worth noting that for SlowFast-R50, the improvement is up to \textbf{4.2} mAP. Moreover, compared with Nonlocal that introduces 10\% overhead and LFB~\cite{lfb} with 182\% overhead, our PGT provides much more performance improvements.

Note that in Tab.~\ref{tab:charades_res}, we report the performances achieved by ``progressive long view" inference method. The performances of ``original long view" inference method are $\sim$3\% lower in mAP, which reveals that ``pg long" method is more suitable for long activities consisting of several sub-actions.

\vspace{-0.07in}
\section{Experiments on AVA}
\vspace{-0.05in}

The AVA~\cite{ava} dataset is designed for action detection, which is labelled with bounding-boxes and action categories for each person in 437 movies. Following standard protocol, we report performance (mAP) on 60 classes.

\textbf{Implementation} We adopt SlowOnly~\cite{slowfast} and SlowFast~\cite{slowfast} as our baselines. Following~\cite{ava,sun2018actor,jiang2018human,slowfast}, we utilize the off-the-shelf human detection results originally adopted by SlowFast as our region proposals. The models are pre-trained on Kinetics-400 or Kinetics-600 following prior works~\cite{x3d,slowfast,lfb}.
For PGT, similar to Charades, we adopt ``pg long" inference method and max-pooling over frames to get multi-classification results.

\textbf{Results} Comparisons with baselines of SlowOnly and SlowFast are shown in Tab.~\ref{tab:ava_res}. It is seen that the proposed PGT provides consistent performance improvements. After adopting the progressive training method, SlowOnly-R50 model achieves a better performance than the much larger SlowOnly-R101 model. Consistent with the experiments on Kinetics and Charades, PGT method provides higher performance improvements than the Nonlocal module with lower computational complexity. Note that compared with the original training method adopting a twice longer input length which introduces twice inference complexity, our PGT version has a better performance (+0.6 mAP vs. 16 $\times$ 8 original training version). This verifies that PGT is an efficient training strategy for processing long videos.

\begin{table}[t]
	\caption{\textbf{Comparison with the SOTA on Charades.} All PGT settings are based on $T\times P \times \tau = 16\times 5\times 8$ and their corresponding baselines are $T \times \tau = 16\times8$.}
	\vspace{-0.1in}
	\renewcommand{\arraystretch}{1.0}
	\centering
	\footnotesize
	\setlength\arrayrulewidth{0.7pt}
	\resizebox{\columnwidth}{!}{
		\begin{tabular}{l|c|c|c|c}
			model & backbone & pretrain & mAP & GFLOPs$\times P \times v$\\
			\hline
			Nonlocal~\cite{nonlocal} & R101 & IN+K400 & 37.5 & 544 $\times$1$\times$30\\
			STRG, +NL~\cite{strg} & R101 & IN+K400 & 39.7 & 630$\times$1$\times$30 \\
			Timeception~\cite{timeception} & R101 & K400 & 41.1 & N/A \\
			LFB, +NL~\cite{lfb} & R101 & K400 & 42.5 &  529$\times$1$\times$30\\
			X3D-XL~\cite{x3d} & - & K400 & 43.4 & 48.4$\times$1$\times$30 \\
			\hline
			\rowcolor[gray]{0.95} SlowOnly~\cite{slowfast} & R50 & K400 &37.3 & 109$\times$1$\times$30\\
			\rowcolor[gray]{0.95} \textbf{SlowOnly, + PGT} & R50 & K400 & 40.3 & 109$\times$5$\times$6 \\
			SlowOnly & R101 & K400 & 39.0 & 187 $\times$1$\times$30 \\
			\textbf{SlowOnly, + PGT} & R101 & K400 & 42.7 & 187$\times$5$\times$ 6\\
			\rowcolor[gray]{0.95} SlowFast~\cite{slowfast} & R50 & K400 & 39.6 & 130$\times$1$\times$30\\
			\rowcolor[gray]{0.95} \textbf{SlowFast, +PGT} & R50 & K400 & 43.8 & 130$\times$5$\times$6\\
			SlowFast~\cite{slowfast} & R101 & K400 & 42.1 & 213$\times$1$\times$30\\
			SlowFast, +NL & R101 & K400 & 42.5 & 234$\times$1$\times$30\\
			\textbf{SlowFast, +PGT} & R101 & K400 & 44.3 & 213$\times$5 $\times$6\\
			\rowcolor[gray]{0.95} SlowFast, +NL & R101 & K600 & 45.2 & 234$\times$1$\times$30\\
			\rowcolor[gray]{0.95} \textbf{SlowFast, +PGT} & R101 & K600 & 47.7 & 213$\times$5$\times$ 6\\
	\end{tabular}}
	\vspace{-0.05in}
	\label{tab:charades_res}  
\end{table}  

\begin{table}[t]
	\caption{\textbf{Performances on AVA-v2.2.} Our PGT variants have progressive step $P=5$.}
	\vspace{-0.1in}
	\renewcommand{\arraystretch}{1.0}
	\centering
	\footnotesize
	\setlength\arrayrulewidth{0.7pt}
	\begin{tabular}{l|p{0.5in}|p{0.5in}}
		model & pretrain & val mAP \\
		\hline
		SlowOnly, R50, 8$\times$8~\cite{slowfast} & K400 & 21.9\\
		SlowOnly, R50, 16$\times$8~\cite{slowfast} & K400 & 22.9\\
		\textbf{SlowOnly, R50, 8$\times$5$\times$8, + PGT} & K400 & 23.5\\
		\rowcolor[gray]{0.95} SlowOnly, R101, 8$\times$8~\cite{slowfast} & K400 & 23.4\\
		\rowcolor[gray]{0.95} \textbf{SlowOnly, R101, 8$\times$5$\times$8, + PGT} & K400 & 24.5\\
		SlowFast, R101, 8$\times$8, + NL~\cite{x3d} & K600 & 27.4\\
		\textbf{SlowFast, R101, 8$\times$5$\times$8, + PGT} & K600 & 27.6\\
		
	\end{tabular} 
	\vspace{-0.15in}
	\label{tab:ava_res}  
\end{table}  

\vspace{-0.03in}
\section{Conclusion}
\vspace{-0.05in}
  We propose the progressive training (PGT) method for end-to-end training models on long videos. With the redesigned Markov convolutional operators, we split an integrated computation progress into several serial progressive steps satisfying one-way dependency, which reduces the resource requirement while ensuring the integrity of temporal semantics. With out-of-box settings, it works on multiple advanced video backbones and benchmarks, achieving 1$\sim$4\% higher performances with negligible computation or parameter overhead. We hope PGT can provide a new option for researchers who need to process long videos.

\vspace{-0.05in}
\paragraph{Acknowledges}
This work is supported in part by the National Key R\&D Program of China, No. 2017YFA0700800, National Natural Science Foundation of China under Grants 61772332 and Shanghai Qi Zhi Institute, SHEITC (018-RGZN-02046).
  
\clearpage
{\small
\bibliographystyle{../style/ieee_fullname}
\bibliography{egbib}
}

\end{document}